\documentclass{article}
\usepackage{spconf,amsmath,graphicx,hyperref}

\usepackage{algorithm}
\usepackage{algorithmic}
\usepackage{amssymb}
\usepackage{amsfonts}       
\usepackage{booktabs}
\usepackage{multirow}
\DeclareMathAlphabet{\mathbbold}{U}{bbold}{m}{n}
\usepackage{makecell}
\usepackage{arydshln} 

\newcommand{\myname}{EQUISeg}

\newcommand{\mone}{CMTB}
\newcommand{\mtwo}{SGM}

\title{Robust Multimodal Semantic Segmentation with Balanced Modality Contributions}
\name{Jiaqi Tan$^{1}$, $^\dagger$Xu Zheng$^{2,3}$, Fangyu Li$^{1}$, Yang Liu$^{1,^\ast}$\thanks{$^\ast$ Corresponding author. Email: yang.liu@bupt.edu.cn \ $^\dagger$ Project Lead.}}
\address{\textsuperscript{\rm 1}Beijing University of Posts and Telecommunications\\
\textsuperscript{\rm 2}HKUST(GZ), \textsuperscript{\rm 3}INSAIT, Sofia University “St. Kliment Ohridski”}
\begin{document}
\ninept
\maketitle
\begin{abstract}
Multimodal semantic segmentation enhances model robustness by exploiting cross-modal complementarities. However, existing methods often suffer from imbalanced modal dependencies, where overall performance degrades significantly once a dominant modality deteriorates in real-world scenarios. Thus, modality balance has become a critical challenge for practical multimodal segmentation. To address this issue, we propose \myname, a multimodal segmentation framework that balances modality contributions through equal encoding of modalities. Built upon a four-stage Cross-modal Transformer Block (\mone), \myname~enables efficient multimodal fusion and hierarchical selection. Furthermore, we design a Self-guided Module (\mtwo) that mitigates modality imbalance by introducing a mutual guidance mechanism, enabling each modality to adaptively adjust its contribution and enhance robustness under degraded conditions. Extensive experiments on multiple datasets demonstrate that \myname~achieves significant performance gains and effectively alleviates the adverse effects of modality imbalance in segmentation tasks.

\end{abstract}
\begin{keywords}
Modality Balance, Multimodal Semantic Segmentation, Model Robustness
\end{keywords}
\section{Introduction}
\label{sec:intro}

With the rise of modular sensors, incorporating more modalities into semantic segmentation has become a mainstream direction for improving the robustness of segmentation models~\cite{chen2020bi,wang2022multimodal,broedermann2023hrfuser,guo2022segnext}. The core idea is to leverage multimodal complementarity, enabling models to remain robust under extreme conditions such as low-light environments at night or sensor failure by exploiting the advantages of modalities unaffected by such conditions.

Existing multimodal semantic segmentation models can be roughly categorized into three types: \textcircled{1} RGB-dominant approaches ~\cite{zhang2023delivering,zhang2023cmx,liu2024fourier}. For example, CMNext~\cite{zhang2023delivering} extracts complementary information from other modalities through its self-query hub design to enhance RGB representations, while CMX~\cite{zhang2023cmx} effectively achieves RGB–modality fusion via FRM and FFM. However, the robustness of these methods heavily relies on the RGB modality. Once RGB becomes fragile under extreme conditions such as low-light or distortion, overall performance drops significantly. \textcircled{2} Adaptive modality selection approaches~\cite{zheng2024centering,zheng2024magic++,zheng2024learning2}. A representative work is MAGIC~\cite{zheng2024centering}, which dynamically selects the dominant modality by calculating the cosine similarity of modality features. Although theoretically sound, this approach often fails in practice due to quality differences across modalities. For instance, high-quality modalities like RGB and Depth tend to retain higher similarity scores even under extreme conditions compared to lower-quality modalities such as Event, causing the adaptive selection mechanism to malfunction. \textcircled{3} Equal-treatment approaches~\cite{zhang2021abmdrnet,li2024stitchfusion,zhu2024customize}. A typical example is StitchFusion~\cite{li2024stitchfusion}, which encodes all modalities equally to obtain modality-specific information, and employs Modality Adapters at different density levels to achieve modality fusion. While these methods show great potential, existing models usually conduct fusion only at the feature level after self-attention encoding and through simple networks such as Feed-Forward Networks~\cite{hu2018squeeze} rather than enabling information interaction during the feature extraction stage, which greatly limits fusion effectiveness.

To address these limitations, we propose \myname, a novel multimodal segmentation framework that extends SegFormer~\cite{xie2021segformer} by integrating the strengths of three mainstream paradigms to achieve balanced representation and robust cross-modal fusion. \myname\ consists of two core modules: the Cross-modal Transformer Block (\mone) and the Self-Guided Module (\mtwo). The encoder is built from four hierarchical \mone s, each adopting a primary–auxiliary scheme where one modality is encoded as primary and others as auxiliaries. This design emphasizes cross-modal complementarity and avoids over-reliance on a single modality, with self-attention capturing intrinsic features and cross-attention injecting informative auxiliary cues directly during representation learning. To further mitigate modality imbalance, \mtwo\ enforces equal contribution by randomized teacher–student guidance after prototype abstraction~\cite{wang2020intra,tan2025robust}, fundamentally reducing fixed modality dependence and enhancing robustness under diverse and degraded sensing conditions.

Extensive experiments on the DeLiVER~\cite{zhang2023delivering}, and MUSES~\cite{brodermann2024muses} datasets under both the mIoU benchmark and robustness evaluation criteria~\cite{liao2025benchmarking} demonstrate that our framework achieves improvements of +1.56\% and +1.85\% compare with SoTA, respectively, outperforming previous methods.


\section{Method}
\label{sec:method}
\begin{figure*}[t!]
\centering
\includegraphics[width=1.0\textwidth]{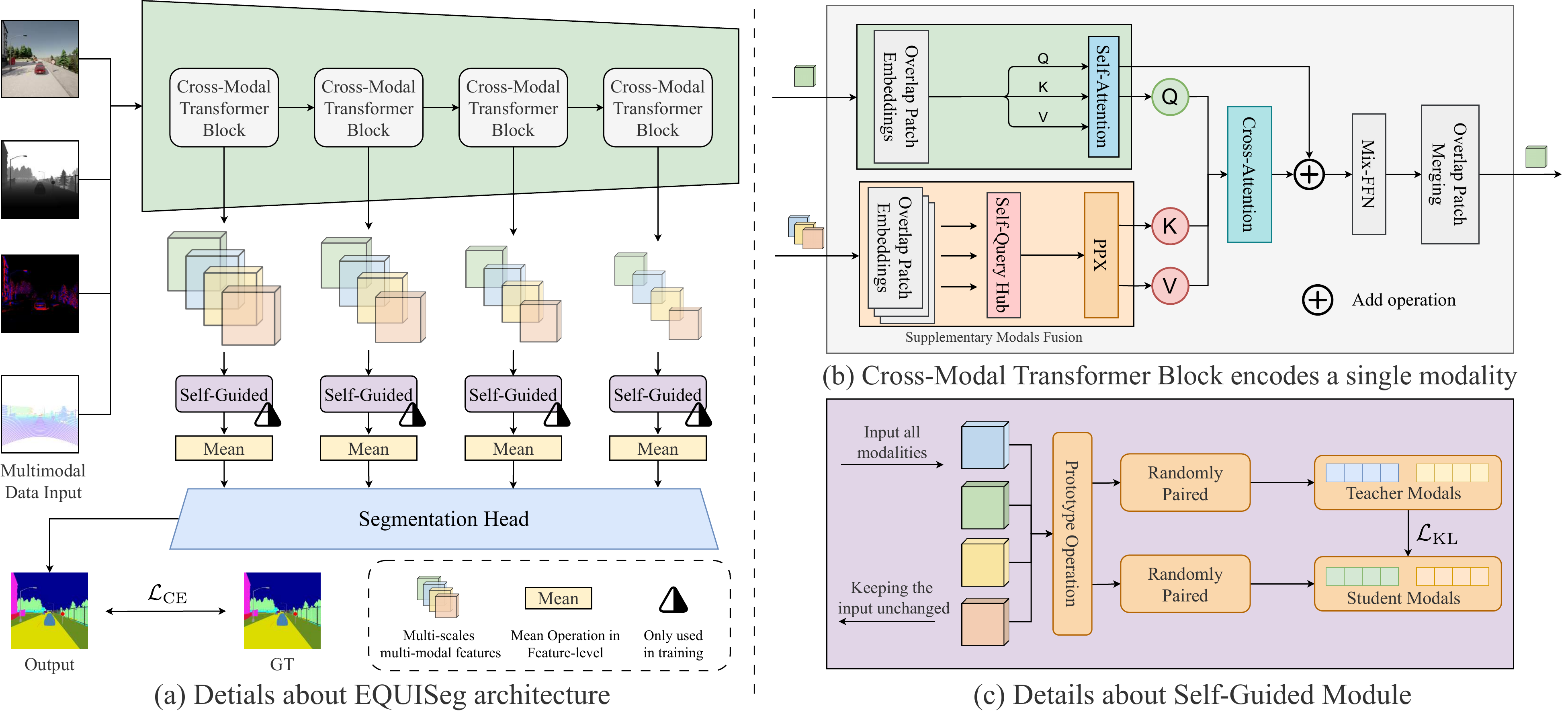}
\vspace{-15pt}
\caption{Overview of the \myname~framework. Figure (a) shows the overall architecture, where multimodal data $\{r, d, e, l\}$ are encoded through multi-layer attention and then fed into the segmentation head. The framework contains two core modules: the Cross-modal Transformer Block (\mone) and the Self-Guided Module (\mtwo), with \mtwo\ only used during training. As shown in Figure (b), \mone\ fuses auxiliary modalities via the Self-Query Hub (SQ Hub) and Parallel Pooling Mixer (PPX)~\cite{zhang2023delivering}, then integrates them with the primary modality through attention to capture complementary features. Figure (c) illustrates \mtwo, which assigns teacher--student roles and performs prototype-based self-guidance~\cite{wang2020intra,tan2025robust}, encouraging equal modality contribution without altering the input or output.}
\label{fig:MMSS}
\vspace{-15pt}
\end{figure*}

To achieve robust and effective multimodal semantic segmentation, the proposed \myname\ framework is constructed with two core modules: the Cross-modal Transformer Block (\mone) and the Self-Guided Module (\mtwo). We will introduce the design of \mone\ in Sec.~\ref{sec:CMTB}, the mechanism of \mtwo\ in Sec.~\ref{sec:SGM}, and the overall architecture of \myname\ in Sec.~\ref{sec:architecture}. The overall architecture is illustrated in Fig.~\ref{fig:MMSS}, where the \mtwo\ is only used during training.

\subsection{Cross-modal Transformer Block}
\label{sec:CMTB}
The detailed structure of the Cross-modal Transformer Block is shown in Fig.~\ref{fig:MMSS} (b). Obtaining pyramid features through multi-layer Transformer Blocks can effectively improve segmentation performance and has been widely adopted in various advanced multimodal frameworks~\cite{zheng2024centering,zheng2024magic++}. However, relying solely on self-attention to process multimodal information is insufficient. Therefore, the core of the Cross-modal Transformer Block is to introduce cross-attention~\cite{guo2022segnext} to enhance modality fusion. Specifically, inspired by the dual-branch design of CMNeXt~\cite{zhang2023delivering} and the equal encoding strategy of StitchFusion~\cite{li2024stitchfusion}, we divide the input \(N\) modalities into a combination of one primary modality and \(N-1\) auxiliary modalities. The \(N-1\) auxiliary modalities are processed by the Self-Query Hub~\cite{zhang2023delivering} and PPX modules~\cite{zhang2023delivering} to select informative auxiliary features, which are then fused with the primary modality via cross-attention. Each modality is encoded independently to obtain rich and equally treated features. For example, when there are four input modalities $\{f_1, f_2, f_3, f_4\}$, encoding modality $f_1$ treats $\{f_2, f_3, f_4\}$ as auxiliary modalities. The whole process can be formulated as follows: $f_1'=\text{MHSA}(f_1), \; f_{\text{aux}}=\text{PPX}(\text{SQ Hub}(f_2,f_3,f_4))$, where $\text{MHSA}$ denotes the multi-head self-attention module in Segformer~\cite{xie2021segformer}.

After obtaining the modality feature \(f_1'\) and the fused auxiliary feature \(f_{aux}\), residual cross-attention based feature fusion is performed, which can be formulated as follows:
\vspace{-6pt}
\begin{gather}
f_{c}=\text{MHCA}(f_1,f_2)= \text{Softmax}\left(\frac{Q_1K_2^T}{\sqrt{d_k}}\right)V_2, \label{eq:MHCA}
\end{gather}
where the query is given by \(Q_1 = f_1' W_Q\), while the key and value are obtained as \(K_2 = f_{aux} W_K\) and \(V_2 = f_{aux} W_V\), respectively. The final output feature is obtained through a residual connection:  
\begin{gather}
f=f_1' + f_c,
\end{gather}
where \(f\) denotes the final modality feature, \(\text{MHCA}\)~\cite{guo2022segnext} represents the multi-head cross-attention mechanism, and \(f_c\) denotes the cross-modal feature obtained via cross-attention. In the EQUISeg architecture, multi-stage MTCBs are assigned to all modalities for balanced encoding, thereby effectively mitigating modality dependency.

\subsection{Self-Guided Module}
\label{sec:SGM}
The detailed structure of the Self-Guided Module is shown in Fig.~\ref{fig:MMSS} (c). At present, the mainstream approach to modality balancing is to design plug-and-play modules that take the features requiring balance as input, compute a loss function, and leverage this loss to guide balanced modality embedding~\cite{gat2020removing,ni2025rollingq,cicchetti2024gramian}. Meanwhile, cross-modal knowledge distillation has achieved remarkable results in modality transfer~\cite{tan2025robust,gu2025breaking}. Therefore, we design a Self-Guided Module (\mtwo) that divides the modalities into teacher and student modalities. After prototypical representation of the features, the module performs modality information balancing through knowledge distillation. To avoid selection failure caused by quality differences among modalities, the grouping process is carried out randomly. Specifically, we firstly perform the following prototyping operation on the feature $f$ with class $C$:
\begin{equation}
    p=[p_0,p_1,\ldots,p_C],\quad p_c=\frac{\sum_j f^{j}\mathbbold{1}[l_j = c]}{\sum_j \mathbbold{1}[l_j = c]},
    \label{eq:proto}
\end{equation}

where $f_{j}$ denotes the feature of pixel $j$, $p$ denotes the prototype feature, and $l_j$ represents the ground truth label of pixel $j$. The indicator function $\mathbbold{1}$ outputs 1 if the specified condition is satisfied and 0 otherwise. Taking DeLiVER as an example, at each training step the prototype features $[p_r, p_d, p_e, p_l]$ are randomly divided into a teacher feature set $[p_{t1}, p_{t2}]$ and a student feature set $[p_{s1}, p_{s2}]$. Modality self-guidance is then performed by minimizing the Kullback–Leibler divergence, defined as  
\(
    \mathcal{L}_{\text{KL}}(f_t,f_s) = \sum_{c=1}^C f_t(c) \log \frac{f_t(c)}{f_s(c)},
\)
where $f_t$ and $f_s$ denote the teacher and student features, respectively. Based on this, the self-guidance loss is computed as  
\begin{gather}
    \mathcal{L}_s = \sum_{i=1}^{4}\big(\mathcal{L}_{\text{KL}}(f_{t1}^i,f_{s1}^i)+\mathcal{L}_{\text{KL}}(f_{t2}^i,f_{s2}^i)\big), \label{eq:Loss_s_c}
\end{gather}
where $i$ denotes the $i$-th Transformer block. When the number of modalities is odd, one modality is randomly ignored in each round before performing the above grouping operation. This interleaved self-guidance strategy effectively prevents modalities from overfitting to high-quality ones, thereby better exploiting the complementary information contained in low-quality modalities. 

\subsection{\myname\ Architecture}
\label{sec:architecture}

As shown in Fig.~\ref{fig:MMSS}(a), the M-Segformer adopts a multi-scale encoder–decoder architecture. The multi-scale encoder consists of a four-stage multi-branch encoder, where the number of branches corresponds to the number of modalities. Each branch specifically encodes one modality, while treating the others as auxiliary modalities that are fused through the Cross-modal Transformer Block (\mone\ in Sec.\ref{sec:CMTB}). The four-stage structure follows prior advanced Transformer models to extract pyramid features~\cite{xie2021segformer,wang2021pyramid}, which are then fed into the segmentation head~\cite{xie2021segformer} for semantic segmentation. Specifically, for the input multimodal data  \(\{r, d, e, l\}\), the operations of \myname\ can be formulated as follows:
\vspace{-6pt}
\begin{equation}
    \{f_r^i,f_d^i,f_e^i,f_l^i,\dots\}_{i=1}^{4} = \text{CMTB}_{i=1,n=1}^{4,N}(\{r, d, e, l\dots\}),
    \label{eq:CMTB}
\end{equation}
where, \(\{r, d, e, l, \dots\}\) denotes the input raw modality data, \(i\) represents the features obtained from the \(i\)-th Transformer block, \(N\) denotes the total number of modalities, and \(n\) is the modality index. For example, in the DeLiVER dataset, \(N=4\), and \(\{f_r^i, f_d^i, f_e^i, f_l^i, \dots\}_{i=1}^{4}\) represents the final multi-stage modality feature outputs. Then, the features are fed into the Self-Guided Module (\mtwo\ in Sec.\ref{sec:SGM}) for modality balancing, yielding the balanced loss \(\mathcal{L}_s\). The formulation of \(\mathcal{L}_s\) is as follows:
\vspace{-6pt}
\begin{equation}
    \mathcal{L}_s = \text{SGM}(\{f_r^i,f_d^i,f_e^i,f_l^i,\dots\}_{i=1}^{4}),
    \label{eq:loss_s}
\end{equation}
The calculation method is as shown in Eq.~\ref{eq:Loss_s_c}. Finally, the features are fed into the segmentation head for segmentation \(y = \text{SegHead}(\{f_r^i,f_d^i,f_e^i,f_l^i,\dots\}_{i=1}^{4})\) , and the cross-entropy loss is obtained as follows:
\vspace{-6pt}
\begin{gather}
    \mathcal{L}_{\text{CE}} = -\sum^{C}_{c=1}g_{c}\text{log}(\text{Softmax}(y_{c})),\label{eq:L_CE}
\end{gather}
where, \(C\) is the number of classes, and \(g\) denotes the ground truth. By combining Eq.~\ref{eq:loss_s} and~\ref{eq:L_CE}, the final loss can be obtained as follows:
\vspace{-6pt}
\begin{equation}
    \mathcal{L} = \mathcal{L}_{\text{CE}} + \lambda\mathcal{L}_s,
    \label{eq:L}
\end{equation}
where, \(\mathcal{L}\) represents the loss during training.  \(\lambda\) is a hyperparameter that controls the balancing strength.

\section{EXPERIMENTS}
\label{sec:EXPERIMENTS}

\begin{table*}[ht!]
\centering
\resizebox{\textwidth}{!}{
\begin{tabular}{c|cccccccc|cccccccc}
\toprule
\multirow{2}{*}{\makecell{Model\\ (\#Params(M))}} & \multicolumn{8}{c|}{DeLiVER} & \multicolumn{8}{c}{MUSES} \\
\cmidrule(lr){2-9} \cmidrule(lr){10-17}
                       & mIoU &  \makecell{$\text{EMM}$\\ $(\text{Avg})$} & $\makecell{$\text{EMM}$\\ $(\text{p=0.1})$}$ & \makecell{$\text{RMM}$\\ $(\text{Avg})$} & \makecell{$\text{RMM}$\\ $(\text{p=0.1})$} & \makecell{$\text{NM}$\\ $(\text{Low})$} & \makecell{$\text{NM}$\\ $(\text{Mid})$} & Mean
                       & mIoU &  \makecell{$\text{EMM}$\\ $(\text{Avg})$} & $\makecell{$\text{EMM}$\\ $(\text{p=0.1})$}$ & \makecell{$\text{RMM}$\\ $(\text{Avg})$} & \makecell{$\text{RMM}$\\ $(\text{p=0.1})$} & \makecell{$\text{NM}$\\ $(\text{Low})$} & \makecell{$\text{NM}$\\ $(\text{Mid})$} & Mean \\
\midrule
CMNeXt(58.73)~\cite{zhang2023delivering} & 66.33 & 37.90 & 60.41 & 47.49 & 62.68 & \textbf{35.23} & 16.37 & 46.63
             & 46.66 & 20.80 & 31.48 & 26.53 & 42.17 & 18.32 & 10.52 & 28.07 \\
MAGIC(24.73)~\cite{zheng2024centering}   & 66.10 & 44.97 & 62.68 & 48.19 & 63.01 & 24.03 & 13.31 & 46.04
             & 49.02 & 33.34 & 42.37 & 37.02 & 46.24 & 13.34 & 8.46 & 32.83 \\
MAGIC++(27.51)~\cite{zheng2024magic++}   & 67.34 & 44.85 & 63.52 & 49.31 & 64.07 & 33.12 & 18.31 & \underline{48.65}
             & 50.14 & 34.52 & 43.84 & 37.34 & 46.90 & 15.43 & 9.15 & \underline{33.90} \\
StitchFusion(26.50)~\cite{li2024stitchfusion} & \textbf{68.20} & 41.98 & 63.29 & 48.33 & 64.53 & 24.91 & 17.11 & 46.91
             & 48.42 & 26.59 & 40.11 & 35.66 & 45.38 & 12.44 & 10.28 & 31.27 \\
\midrule
\myname(52.77)(Ours) & 67.90 & \textbf{48.22} & \textbf{65.75} & \textbf{50.96} & \textbf{64.64} & 34.87 & \textbf{19.13} & \textbf{50.21}
             & \textbf{50.26} & \textbf{35.63} & \textbf{45.06} & \textbf{38.61} & \textbf{47.63} & \textbf{20.47} & \textbf{12.62} & \textbf{35.75} \\
\bottomrule
\end{tabular} }
\vspace{-6pt}
\caption{Quantitative comparison of multimodal segmentation results on the DeLiVER and MUSES datasets. mIoU(\%) denotes the mean Intersection over Union (IoU) across different cases in the dataset. Entire-Missing Modality (EMM), Random-Missing Modality (RMM), and Noisy Modality (NM)~\cite{liao2025benchmarking} correspond to random removal, block-wise removal, and noise addition to modalities, respectively. The Mean column is the average of all metrics in each dataset, which we take as the \textit{robustness score} to evaluate the overall robustness of each model.}
\label{Tab:Quantitative}
\vspace{-10pt}
\end{table*}

\subsection{Experiment Setup}
\label{sec:setup}
\noindent\textbf{Dataset} \textbf{DeLiVER}~\cite{zhang2023delivering} consists of four modalities (RGB, Depth, LiDAR, Event), covering diverse weather and challenging conditions with 25 semantic categories, and serves as a key benchmark for autonomous driving robustness. \textbf{MUSES}~\cite{brodermann2024muses} provides four modalities (RGB, MEMS LiDAR, FMCW radar, Event), designed for driving under high uncertainty. It includes 2,500 samples across various weather and illumination conditions with 19 semantic categories, each annotated with high-quality 2D panoramic labels. Both datasets cover a wide range of environments and sensor failure cases, enabling comprehensive evaluation under challenging scenarios.

\noindent\textbf{Baseline} We compare \myname\ with advanced models from the three mainstream segmentation architectures mentioned in the introduction, namely CMNeXt~\cite{zhang2023delivering}, MAGIC~\cite{zheng2024centering}, MAGIC++~\cite{zheng2024magic++}, and StitchFusion~\cite{li2024stitchfusion}. CMNeXt~\cite{zhang2023delivering} employs a dual-branch asymmetric encoder structure and serves as a representative of RGB-dominant methods. MAGIC~\cite{zheng2024centering} and MAGIC++~\cite{zheng2024magic++} adopt cosine similarity to select encoding modalities, representing approaches based on adaptive modality selection. StitchFusion~\cite{li2024stitchfusion}, on the other hand, is an advanced method for equal-modality encoding.

\noindent\textbf{Evaluation Metric} We first evaluate segmentation mIoU across different datasets to assess model performance under extreme weather and complex conditions. Subsequently, we employ the robustness benchmarks~\cite{liao2025benchmarking} to assess the model’s performance when sensor information is missing or degraded. This benchmark simulates extreme conditions by applying operations such as complete modality removal, random modality loss, and noise injection, covering three test types: Entire-Missing Modality (EMM), Random-Missing Modality (RMM), and Noisy Modality (NM).

\noindent\textbf{Implementation Details} Experiments on DeLiVER is conducted using 4 NVIDIA A6000 GPUs, while MUSES is trained on 4 NVIDIA RTX 3090 GPUs. We adopt the AdamW optimizer with a learning rate of $6 \times 10^{-5}$, a 10 epoch warm-up, and polynomial decay (exponent 0.9) over 200 epochs. The batch size is set to 2 per GPU, and input images are cropped to $1024 \times 1024$. The model was trained for 12 hours on the Deliver dataset with MIT-B2 as the backbone, and for 7 hours on the MUSES dataset with MIT-B0.

\subsection{Comparison with State of the Art Methods}
\label{sec:result}

\begin{figure}[t!]
\centering
\includegraphics[width=0.48\textwidth]{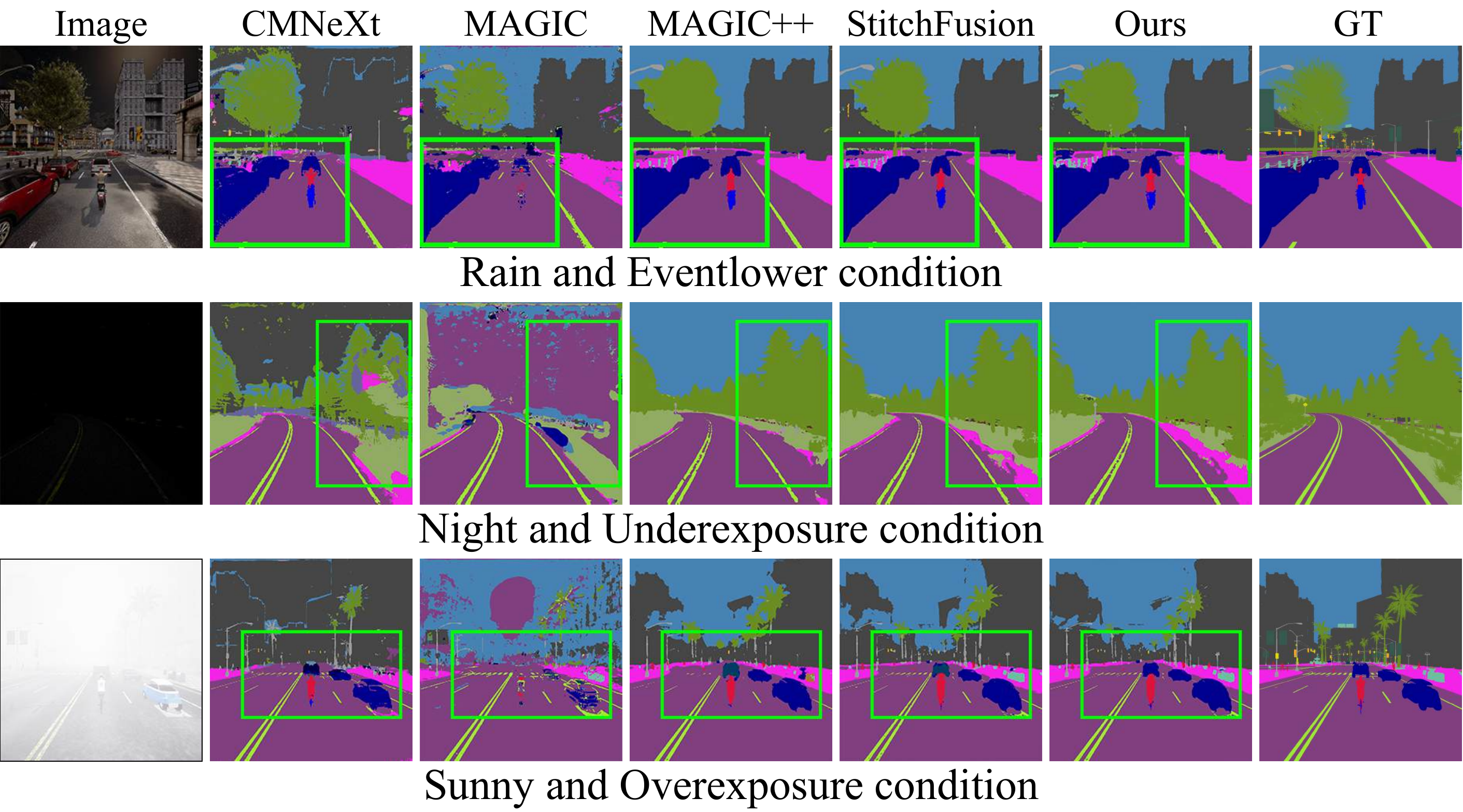}
\vspace{-15pt}
\caption{Qualitative Comparison Results on DeLiVER Datasets.}
\label{fig:Result}
\vspace{-15pt}
\end{figure}

To validate the effectiveness of our proposed \myname\ framework, we conducted experiments on the DeLiVER and MUSES datasets. The quantitative results are presented in Tab.~\ref{Tab:Quantitative}, and the qualitative results are shown in Fig.~\ref{fig:Result}. All models use \textbf{MIT-B2} as the backbone.

\noindent\textbf{Quantitative Results on DeLiVER} On the DeLiVER dataset, our method achieves an mIoU of 67.90, close to StitchFusion (68.20). Although the average accuracy under clean conditions is slightly lower, our model shows clear advantages in extreme cases: for example, under overexposure and underexposure, it achieves 66.70 and 65.61, surpassing StitchFusion (65.49 / 64.18). Compared with CMNeXt, which heavily relies on RGB and thus collapses under modality absence, our method achieves much higher scores in the robustness benchmarks. Specifically, we obtain the best results in EMM (48.22 (Avg), 65.75 (p=0.1)) and RMM (50.96 (Avg), 64.64 (p=0.1)). StitchFusion, though strong in clean mIoU, drops more severely in these tests, while our method maintains balanced performance. For noisy modalities, our approach performs best at mid-level noise (19.13) and remains competitive under low noise (34.87 vs CMNeXt’s 35.23). Overall, our model reaches the highest robustness score (50.21), outperforming MAGIC++ by +2.56\%, demonstrating stronger resilience to modality missing and corruption.

\noindent\textbf{Qualitative Results on DeLiVER} Fig.~\ref{fig:Result} further confirms these advantages. Under rain and eventlower, our model produces more coherent pedestrians; under extreme RGB degradations (night and overexposure), it better preserves road boundaries, vegetation, and object contours. These results demonstrate that our method achieves both high accuracy and strong robustness in challenging conditions.

\noindent\textbf{Quantitative Results on MUSES} On the MUSES dataset, our method achieves the best mIoU of 50.26, slightly higher than MAGIC++ (50.14) and clearly above CMNeXt (46.66). Under robustness benchmarks, our model consistently outperforms baselines: 35.63 / 45.06 (EMM-Avg / p=0.1) and 38.61 / 47.63 (RMM-Avg / p=0.1), all ranking first. For noisy modalities, we also obtain the best results (20.47 / 12.62 under low/mid noise). Overall, our approach achieves the highest robustness score of 35.75, exceeding MAGIC++ by +1.85\%, demonstrating strong resilience to modality absence and noise corruption in extreme conditions.

\subsection{Ablation Study}
\label{sec:ablation}

\begin{table}[t!]
\centering
\resizebox{0.48\textwidth}{!}{
\begin{tabular}{l|c|c}
\hline
Structure & mIoU(\%) & EMM(Avg) \\
\hline
\mone                           & 63.26         & 44.77         \\ \hdashline
with Mean instead Self-Query Hub        & 62.57 (-0.69) & 42.83 (-1.94) \\
without Cross Attention         & 61.92 (-1.34) & 39.36 (-5.41) \\
without Add operation           & 58.69 (-4.57) & 38.61 (-6.16) \\
\hline
\end{tabular}}
\vspace{-6pt}
\caption{Ablation study of the \mone\ architecture. We use MIT-B0 as the backbone and evaluate the mIoU and EMM(Avg) metrics.}
\label{tab:ab-cmtb}
\vspace{-10pt}
\end{table}

\noindent\textbf{Influence of \mone} From Tab.~\ref{tab:ab-cmtb}, we can observe that removing or replacing the core components of \mone\ leads to a clear performance drop in both segmentation accuracy (mIoU) and robustness under missing modalities (EMM). Specifically, replacing the Self-Query Hub with a simple mean operation only slightly decreases mIoU (-0.69) but causes a larger decline in EMM (-1.94), indicating that the hub mainly contributes to robust feature alignment when modalities are incomplete. Removing the cross-attention module results in a more significant drop in both mIoU (-1.34) and EMM (-5.41), showing that multi-modal interaction is crucial for exploiting complementary information across modalities and maintaining robustness under modality absence. The most severe degradation occurs when the Add operation is removed, where mIoU decreases by -4.57 and EMM by -6.16. This highlights that residual fusion through the Add operation is fundamental for effective feature integration and stable optimization. Overall, these results demonstrate that the Self-Query Hub enhances robustness, the cross-attention provides cross-modal complementarity, and the Add operation plays a central role in ensuring stable and effective multimodal learning.

\begin{table}[t!]
\centering
\resizebox{0.48\textwidth}{!}{
\begin{tabular}{l|cccc|c}
\hline
Structure & RGB & Depth & Event & Lidar & EMM(Avg) \\
\hline
\myname\ + \mtwo\ (\(\lambda=60\)) & 81.71 & 89.72 & 29.82 & 32.24 & 44.77 \\ \hdashline
without \mtwo & 88.14 & 93.45 & 8.25  & 11.35 & 41.54 (-3.23) \\
without Prototype Operation & 72.22 & 80.36 & 1.53  & 2.16  & 39.84 (-4.93) \\
\makecell[l]{with Cosine Similarity \\ instead Randomly Paired}  & 82.03 & 91.56 & 22.66 & 24.91 & 43.11 (-1.66) \\
\hdashline
\myname\ + \mtwo\ (\(\lambda=1\)) & 85.22 & 91.64 & 22.47 & 26.75 & 43.46 (-1.31) \\
\myname\ + \mtwo\ (\(\lambda=40\)) & 83.02 & 90.54 & 28.11 & 30.46 & 44.39 (-0.38) \\
\myname\ + \mtwo\ (\(\lambda=80\)) & 78.56 & 87.41 & 31.21 & 32.43 & 44.53 (-0.24) \\
\hline
\end{tabular}}
\vspace{-6pt}
\caption{Ablation study of the \mtwo\ module. We use MIT-B0 as the backbone and evaluate the mIoU metrics. The middle four modality metrics represent the cosine similarity between each modality feature and the fused feature.}
\label{tab:ab-sgm}
\vspace{-10pt}
\end{table}

\noindent\textbf{Effective of \mtwo} As shown in Tab.~\ref{tab:ab-sgm}, enabling \mtwo\ ($\lambda=60$) yields a balanced similarity profile between each modality and the fused feature (RGB/Depth 81.71/89.72; Event/Lidar 29.82/32.24) with the best EMM of 44.77. Removing \mtwo\ collapses Event/Lidar similarities to 8.25/11.35 and reduces EMM by 3.23, indicating that weak modalities are ignored. Removing the prototype operation further drops them to 1.53/2.16 and lowers EMM by 4.93, highlighting the importance of class prototypes for robust distillation. Replacing random teacher--student grouping with cosine-similarity pairing yields lower Event/Lidar (22.66/24.91) and $-1.66$ EMM, suggesting that overly similar pairs weaken the distillation signal. Overall, \mtwo\ plays a significant role in modality balancing, where prototype-based features and random pairing are the key mechanisms for exploiting the value of weak modalities.

\noindent\textbf{Influence of Hyperparameter} Sensitivity analysis shows that the hyperparameter $\lambda$ in Eq.~\ref{eq:L} is crucial for balancing strong and weak modalities. As shown in Tab.~\ref{tab:ab-sgm}, a small value ($\lambda=1$) under-regularizes the process (Event/Lidar 22.47/26.75; $-1.31$ EMM), while a large value ($\lambda=80$) strengthens weak modalities (31.21/32.43) at the expense of RGB/Depth (78.56/87.41), yielding only marginal robustness gains (EMM 44.53). Setting $\lambda=60$ achieves the best trade-off, avoiding overfitting to strong modalities while preventing excessive assimilation of weak modalities.

\section{Conclusion}

In this paper, we address modality bias in multimodal semantic segmentation and propose \myname, an arbitrary multi-modal segmentation framework. \myname\ employs a four-stage Cross-modal Transformer Block (CMTB) for early, efficient interaction and hierarchical selection in representation learning. Self-Guided Module (SGM), active only during training, reduces overfitting to dominant modalities via random teacher–student guidance and prototype distillation, ensuring robustness under modality degradation or absence. Experiments show \myname\ matches or surpasses state-of-the-art results on DeLiVER and MUSES in mIoU and robustness (EMM/RMM/NM). Ablations confirm the role of CMTB’s interaction and residual fusion, and SGM’s prototypical and random pairing strategies in mitigating modality imbalance.

\newpage
\bibliographystyle{IEEEbib}
\bibliography{strings,refs}

\end{document}